\def\year{2020}\relax
\documentclass[letterpaper]{article} 
\usepackage{aaai20}  
\usepackage{times}  
\usepackage{helvet} 
\usepackage{courier}  
\usepackage[hyphens]{url}  
\usepackage{graphicx} 
\usepackage{graphicx}  
\usepackage{amsfonts}
\usepackage{amsmath}
\usepackage{booktabs}

\title{Not All Attention Is Needed: Gated Attention Network for Sequence Data}
\author{Lanqing Xue,\textsuperscript{\rm 1}
		Xiaopeng Li,\textsuperscript{\rm 2\thanks{Work was done prior to joining Amazon.}} 
		Nevin L. Zhang\textsuperscript{\rm 1,3}\\ 
\textsuperscript{\rm 1}The Hong Kong University of Science and Technology, Hong Kong\\
\textsuperscript{\rm 2}Amazon Web Services, WA, USA\\
\textsuperscript{\rm 3}HKUST-Xiaoi Joint Lab, Hong Kong\\ 
lxueaa@cse.ust.hk,xiaopel@amazon.com,lzhang@cse.ust.hk
}
\begin{document}

\maketitle

\begin{abstract}
Although deep neural networks generally have fixed network structures, the concept of dynamic mechanism has drawn more and more attention in recent years. Attention mechanisms compute input-dependent dynamic attention weights for aggregating a sequence of hidden states. Dynamic network configuration in convolutional neural networks (CNNs) selectively activates only part of the network at a time for different inputs. In this paper, we combine the two dynamic mechanisms for text classification tasks.
Traditional attention mechanisms attend to the whole sequence of hidden states for an input sentence, while in most cases not all attention is needed especially for long sequences.
We propose a novel method called Gated Attention Network (GA-Net) to dynamically select a subset of elements to attend to using an auxiliary network, and compute attention weights to aggregate the selected elements. It avoids a significant amount of unnecessary computation on unattended elements, and allows the model to pay attention to important parts of the sequence. Experiments in various datasets show that the proposed method achieves better performance compared with all baseline models with global or local attention while requiring less computation and achieving better interpretability. It is also promising to extend the idea to more complex attention-based models, such as transformers and seq-to-seq models.
\end{abstract}

\section{Introduction}





In recent years, deep learning has achieved great success in many applications, such as computer vision and natural language processing. Various neural network structures have been proposed to solve challenging problems. Although deep neural networks generally have fixed network structures, the concept of dynamic mechanism has drawn more and more attention. Instead of having a fixed computational graph, neural networks with dynamic mechanism adaptively determine how the computation should be conducted based on the inputs.

Attention mechanism is one of such dynamic mechanisms with dynamic weights. Motivated by human visual attention, attention mechanism computes input-dependent dynamic attention weights to select a portion of the input to pay attention to in a soft manner. In image captioning, attention mechanism allows the model to learn alignment between the visual portion of an image and the corresponding word in its text description \cite{xu2015show}. In neural machine translation, an encoder computes a sequence of hidden states from an arbitrary-length sentence, and the decoder needs to extract relevant information from the encoder in order to make predictions of each word. Attention mechanism aggregates the whole sequence of hidden states in the encoder by taking the weighted average of them with attention weights computed according to current decoding context. In such a manner, the decoding of different words in the target sentence pays attention to different words in the source sentence \cite{bahdanau2014neural,vaswani2017attention}. And it has achieved remarkable performance in those applications.

Dynamic network configuration is another dynamic mechanism with dynamic connections for convolutional neural networks (CNNs), and has drawn more and more attention recently. Different from attention mechanism, it selectively activates only part of the network at a time in an input-dependent fashion \cite{bengio2016conditional,zhou2019gaternet}. For example, if we think we a€™re looking at a car, we only need to compute the activations of the vehicle detecting units, not of all features that a network could possible compute \cite{bengio2016conditional}. The benefit of including only part of units for each input is that the propagation through the network will be faster both at training and test time since redundant computations are avoided while the cost of deciding which units to turn on and off is not high. While several works on dynamic network configuration have been proposed for CNNs, such as \cite{bengio2016conditional,veit2018convolutional,zhou2019gaternet}, few such attempts have been made in sequence models such as recurrent neural networks (RNNs) for natural language processing.

In this paper, we seek to combine the two dynamic mechanisms for text classification tasks, and improve attention mechanism by dynamically adjusting attention connections in attention networks. Although attention-based neural networks achieved promising results, common attention mechanism has its limitations. Traditional attention mechanism is generally global, and it attends to all the words in the sentence though some attention weights might be small. However, through investigations into several natural language processing tasks, we observed that only a small part of inputs is related to output targets. It also aligns with our intuition that not all attention is needed especially for long sequences. The computation of attention weights on unrelated elements is redundant. 

Not only that, since attention mechanism assigns a weight to each input unit and even an unrelated unit has a small weight, the attention weights on related units become much smaller especially for long sequences, leading to degraded performance.

\begin{figure}[t]
\centering
\includegraphics[width=\columnwidth]{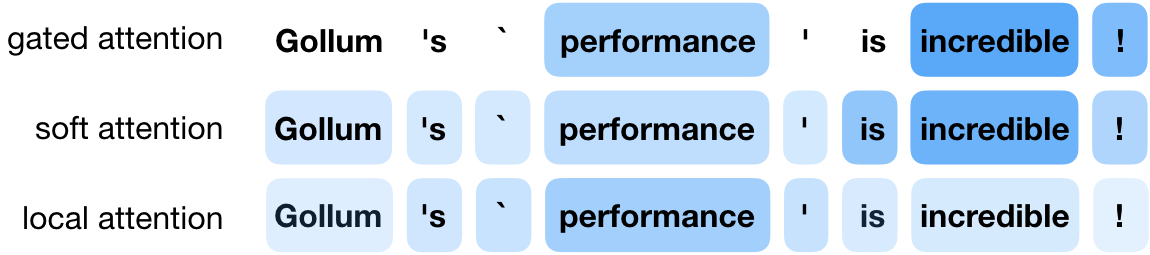} 
\caption{Examples of attention weights indicated by blue color from the proposed GA-Net and attention networks with global and local attentions for sentiment classification. GA-Net has a sparse attention structure and only attends to important words.}
\label{fig1}
\end{figure}
To resolve the problem, we propose a novel method called Gated Attention Network (GA-Net) to dynamically select a subset of elements to attend to using an auxiliary network, and compute attention weights to aggregate the selected elements. A GA-Net contains an auxiliary network and a backbone attention network. The auxiliary network takes a glimpse of the input sentence, and generates a set of input-dependent binary gates to determine whether each word should be paid attention to. The backbone attention network is a regular attention network that performs the major recurrent computation, but only computes attention weights to aggregate the hidden states for the selected words. The attention units are sparsely connected to the sequence of hidden states, instead of densely connected as that in traditional attention mechanism. As an example in Figure \ref{fig1}, the proposed GA-Net has a sparse attention structure and has learned to only attend to important words. The auxiliary network and backbone attention network are trained jointly in an end-to-end manner. In summary, our contributions are as follows:
\begin{itemize}
\item We propose GA-Net, a novel sparse attention network, to dynamically select a subset of elements to attend to using an auxiliary network. It avoids unnecessary attention computation, and allows the model to focus on important elements in the sequence.
\item An efficient end-to-end learning method using gumbel-softmax is proposed to relax the binary gates and enable backpropagation.
\item We conduct experiments on several text classification tasks, and achieve better performance compared with all baseline models in our experiments with global and local attention networks.
\end{itemize}

\subsection{Related Works}
\noindent There have been a lot of works utilizing traditional attention mechanism together with CNNs and RNNs for various applications including computer vision \cite{xu2015show}, speech recognition ~\cite{chorowski2015attention} and natural language processing \cite{radford2018improving,devlin2019bert,cho2015describing,hermann2015teach,rush2015neural,lu2016hierarchical,yang2016stacked,zhou2016attention}.
Attention mechanism has also been extended to act as a sequence model in place of RNNs, such as the Transformer network.
Besides the traditional global structure of attention mechanism, recently, several works have been proposed to make adjustments on the attention mechanism, including inducing task-oriented structure biases to attentions~\cite{kim2017structured,liu2018learning,zhu2017structured,niculae2018sparse}, combining variational approaches with attentions~\cite{deng2018latent}, sparsifing attentions by 
handcrafted attention structures~\cite{guo2019star,ye2019segtree,child2019generating,luong2015effective}, introducing variations of softmax regularizers~\cite{martins2016from,niculae2017regularized,mensch2018differentiable,niculae2018sparse}, and utilizing probabilistic attention with marginalized average method~\cite{yuan2019marginalized}. However, none of these uses an auxiliary network to achieve input-dependent dynamic sparse attention structure.

Several works on dynamic network configuration have been seen in recent years. It is also similar to conditional computation \cite{bengio2013deep}. Gating modules are generally introduced to generate binary gates and dynamically activate part of the network for processing \cite{bengio2016conditional,veit2018convolutional,zhou2019gaternet,denoyer2014deep}. Policy gradient methods or relaxed gating methods are needed in order to enable backpropagation and end-to-end learning. However, these methods are designed for CNNs, and few attempts have been made for sequence models.

\section{Attention Networks}

\begin{figure}[t]
\centering
\includegraphics[width=0.65\columnwidth]{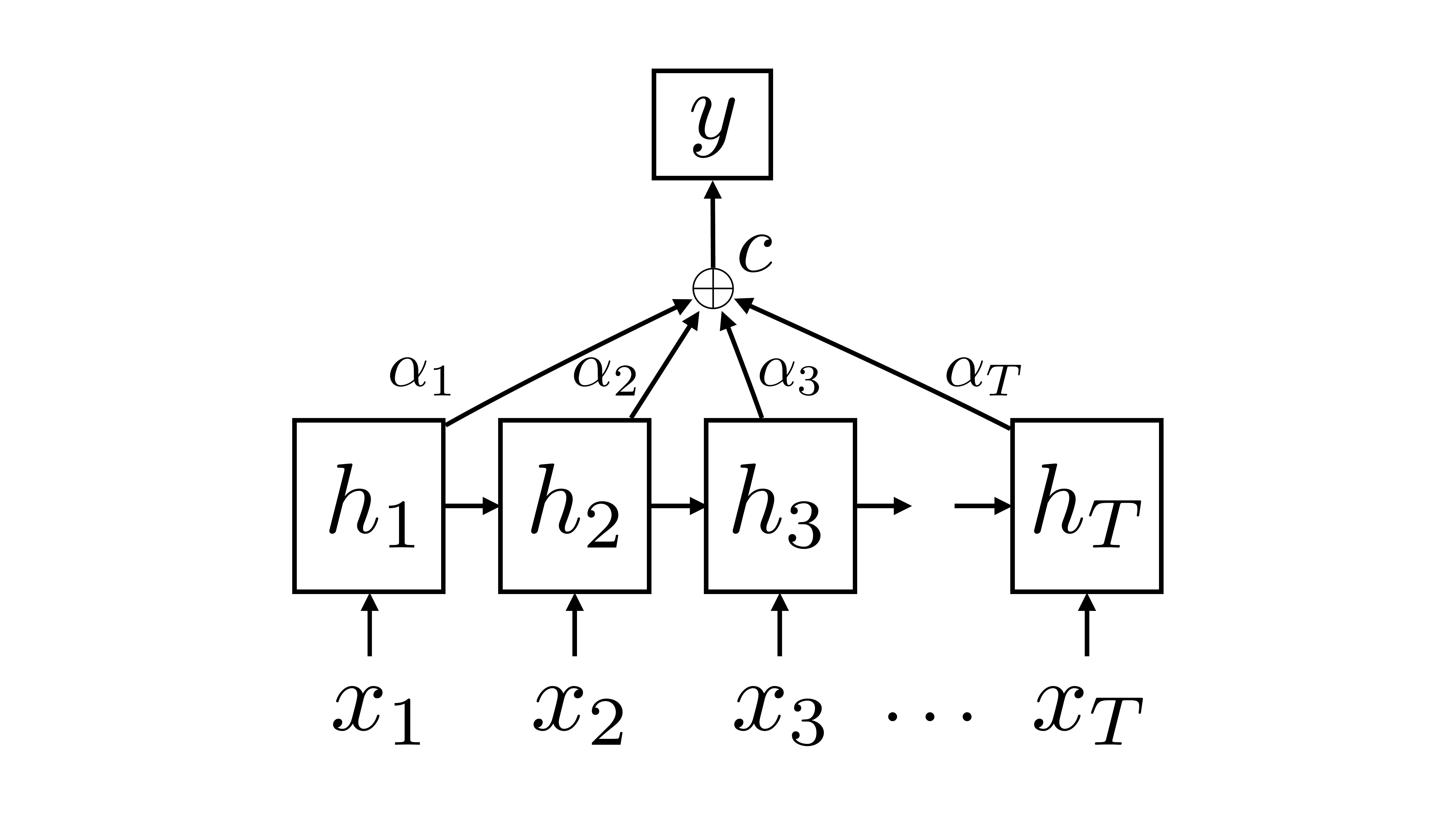} 
\caption{An example of attention mechanism: $\mathbf{c}$ is attention output, and $\alpha$ is attention weight.}
\label{fig2}
\end{figure}

\noindent In natural language processing (NLP), RNNs compute a sequence of hidden states for an arbitrary-length sentence. Instead of requiring the last hidden state to contain all information of the sentence, attention mechanism is widely used to aggregate information from the sequence of hidden states in an input-dependent manner. 
It computes an attention weight for each position in the input source, and takes a weighted average of the hidden states as the output. 
For example, Figure~\ref{fig2} is an LSTM recurrent neural network.  
Let $\mathbf{x}=[\mathbf{x}_1,\mathbf{x}_2,\dots,\mathbf{x}_T]$ be a sequence of inputs, and $\mathbf{h}=[\mathbf{h}_1,\mathbf{h}_2,\dots,\mathbf{h}_T]$ be a sequence of hidden states generated by the LSTM.
The output $\mathbf{y}$ of the model is predicted as follows:
\begin{equation*}
	\hat{\mathbf{y}} = f(\mathbf{c}), 
	\quad 
	\mathbf{c} = \sum_{t=1}^T\alpha_t\mathbf{h}_t,
	\quad
	\mathbf{h_t}=LSTM(\mathbf{h_{t-1}},\mathbf{x}_t)
\end{equation*}
where $\alpha_t$ is attention weight, and it indicates to what extent the $t$-th state influences the output $\mathbf{y}$. Vector
$\mathbf{c}$ is the output of attention unit, and it is a weighted average over all states. 
Attention weight $\alpha_t$ is learned through a Multilayer perceptron (MLP) and normalized over all time steps via a softmax function:
\begin{equation}
\begin{aligned}
	&e_t = MLP(\mathbf{h}_t), \\
	\alpha_t &= softmax_t(\mathbf{e}), \quad\sum_{t=1}^T\alpha_t = 1.
\end{aligned}
\end{equation}
The softmax function makes the sum of attention weights to be $1$. 
Therefore, attention weight can also be seen as probabilities that a state is related to the output. 

Since each state is assigned a weight, this kind of attention mechanism is called soft-attention and global-attention. 
There is also hard-attention and local-attention~\cite{luong2015effective}. In hard-attention, targets attend to only one state each time. 
This is computational efficient. 
However, it loses much information from inputs. 
Usually, not only one input has effects on targets. 
Local attention is a balance between soft and hard attention.
Targets only attend to a window of its neighbors. 
The limitation is obvious, non-neighbors can also have influences on targets. 
Therefore, we proposed a sparse attention mechanism, GA-Net, which can dynamically select important inputs to attend to. 
It not only improves the computational efficiency of soft-attention, but also retains more information and achieves better interpretability than hard-attention and local-attention.

\section{Gated Attention Network}
\begin{figure}[t]
\centering
\includegraphics[width=0.9\columnwidth]{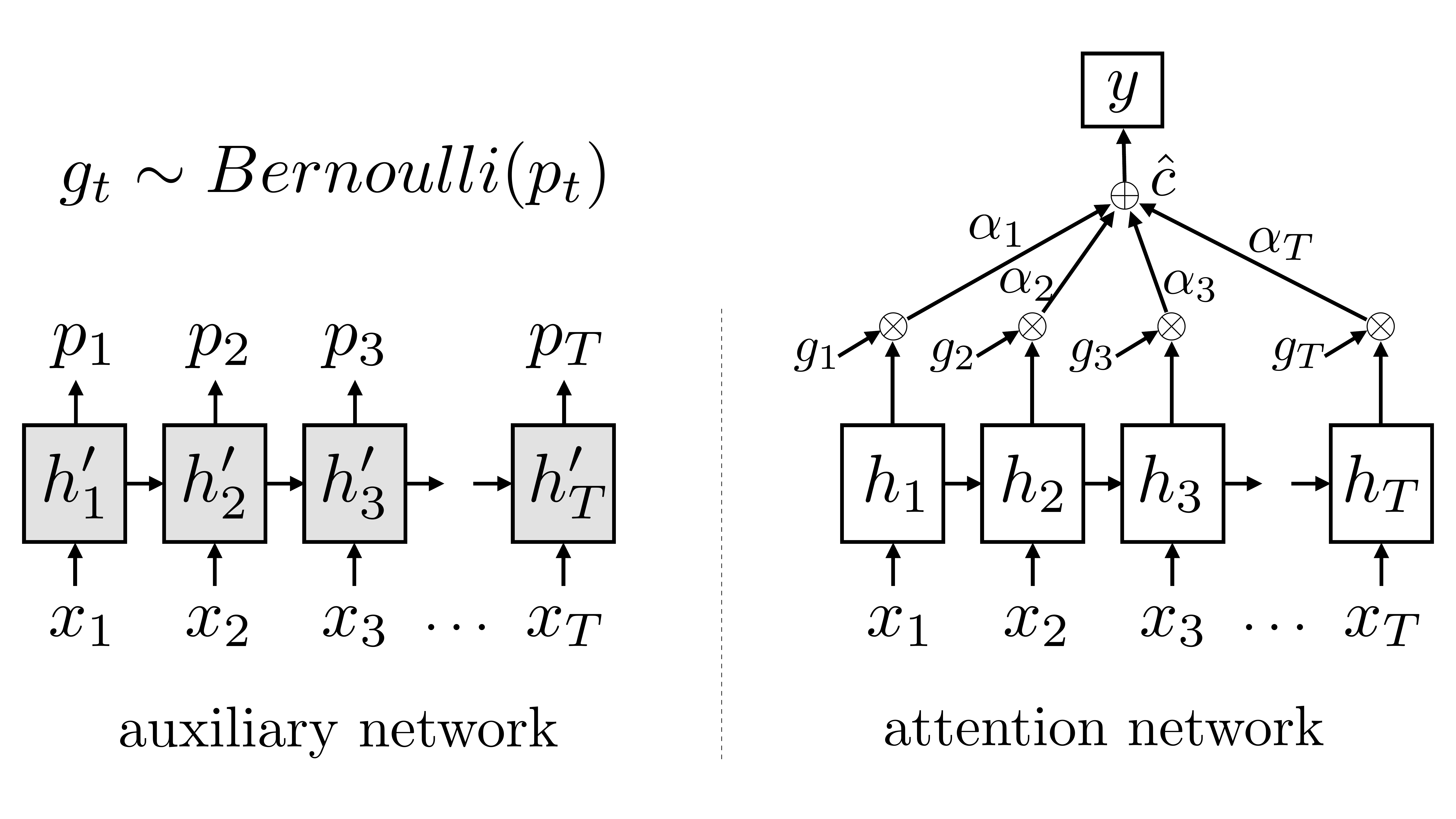} 
\caption{Architecture of GA-Net for classification tasks. Backbone attention network is on the right. The left one is a small auxiliary network producing a series of probabilities. Gate $g_t$ is binary. It is sampled from the output of auxiliary network. Gate is open when $g_t=1$, otherwise closed.}
\label{fig3}
\end{figure}

We call our model Gated Attention Network (GA-Net) because it has an auxiliary network to generate binary gates to dynamically select elements to pay attention to for a backbone attention network. 
Theoretically, the backbone attention network can be any neural network with attention mechanism, such as bidirectional LSTMs and other sequence-to-sequence models.
There are also a range of choices of auxiliary network, as long as it can produce a series of probabilities.
Next, we describe the mechanism of GA-Net in details in the context of text classification.

\subsection{Architecture of GA-Net} 
Figure~\ref{fig3} is the architecture of an GA-Net in classification tasks. 
The backbone attention network on the right is similar to that in Figure~\ref{fig2}. 
The input to the model is a sequence of features $\mathbf{x}=[\mathbf{x}_1,\mathbf{x}_2,\dots,\mathbf{x}_T]$ and the output is the classification target $\mathbf{y}$.
Different from traditional attention network, the backbone network has additional gates $\mathcal{G}=\{g_1,g_2,\dots,g_T\}$, $g_t\in\{0,1\}$ associated with each time step. 
The $t$-th gate is open when $g_t=1$ and is closed when $g_t=0$.
It controls whether the information from current state should flow into targets.
Let $S$ be the set of positions $t$ where $g_{t}=1$. The attention weights in GA-Net are nonzeros for those positions with open gates:
\begin{equation}
\begin{aligned}
	&e_{t} = MLP(\mathbf{h}_{t}) \quad \text{for $t\in S$},\\
	\alpha_{t} &= \frac{\exp{(e_t)}}{\sum_{t'\in S} \exp{(e_{t'})}}, \quad\sum_{t\in S}\alpha_{t} = 1.
\end{aligned}
\end{equation}
On the other hand, for the positions with closed gates, we have $\alpha_t=0$ and those hidden states $\mathbf{h}_t$ are not included into the aggregation.
The attention output and the prediction of $\mathbf{y}$ are then computed as follows:
\begin{equation}
	\hat{\mathbf{y}}=f(\hat{\mathbf{c}}), 
	\quad 
	\hat{\mathbf{c}}=\sum_{t\in S}\alpha_{t}\mathbf{h}_{t}.
\end{equation}

The binary gates act as controllers to selectively activate part of the network. This resembles the dynamic network configuration mentioned earlier. To generate the binary gates, we introduce a dedicated auxiliary network.  The auxiliary network takes a glimpse of the input sentence, and generate the binary gates for each position to determine whether the position needs to be paid attention to. The auxiliary network is on the left of Figure~\ref{fig3}. It shares same input features with backbone attention network, but generally has a much smaller network size.
The output of this auxiliary network is a set of probabilities $\mathbf{p}=\{p_1,p_2,\dots,p_T\}$:
\begin{equation}
\begin{aligned}
	\mathbf{h}_t'&=LSTM(\mathbf{h}_{t-1}', \mathbf{x}_t) \\
	p_t &= sigmoid(\mathbf{Uh}_t').
\end{aligned}
\end{equation}
The probability $p_t$ determines the probability of the gate being open, and it is used to parameterize a Bernoulli distribution. A binary gate is then a sample generated from the Bernoulli distribution:
\begin{equation}
	g_t \sim Bernoulli(p_t).
\end{equation} 
Though the example in Figure~\ref{fig3} shows an LSTM network as the auxiliary network, there are also other choices, such as a feed forward network and a self-attention network. However, a sequence model considering the dependencies among words in the sentence, such as an RNN or a self-attention network might be a better choice in this situation.

\subsection{Training GA-Net}
\noindent We describe the end-to-end training method for the proposed GA-Net in the following.
It is hard to train because the gates have discrete values of $0$ and $1$. 
Thus, errors cannot be back-propagated through gradient descent. 
Some papers~\cite{mnih2014recurrent,luong2015effective} utilized reinforcement learning to solve this problem. 
However, it is computational expensive and has limited performance. 
A recently proposed categorical reparameterization, named Gumbel-Softmax~\cite{jang2017categorical,maddison2017concrete}, is a potential method. 
Gumbel-Softmax aims at approximating a categorical distribution by a Gumbel-Softmax distribution with continuous relaxation. 
Optimizing an objective over an architecture with discrete stochastic nodes can be accomplished by gradient descent on the samples of the corresponding Gumbel-Softmax relaxation. 
Thus, it is a potential silver bullet to the back-propagation problem in our model.

In our model, each gate is a sample of value $0$ or $1$ from Bernoulli distribution. 
It can be taken as a binary `classifier'.
Each classifier $g_t$ produce a two-element one-hot vector $\mathbf{g}_t=[g_{t,i}]_{i=0,1}$, where $g_{t,i}=1$ means $g_t=i$. Similarly, $p_{t,i}$ is the probability that $g_t=i$:
\begin{equation}
	\mathbf{g}_t = \mathrm{one\_hot} \left(\underset{i}{\arg\max} \ p_{t,i}, 
	\ i = 0, 1\right),
\end{equation}
\begin{equation*}
	p_{t,0}=1-p_t,\ p_{t,1}=p_t.
\end{equation*}
To let the auxiliary network differentiable during training, we can apply Gumbel-Softmax distribution as a surrogate of Bernoulli distribution to each gate. The Gumbel-Softmax distribution makes a softmax approximation to the one-hot vector $\mathbf{g}_t$:
\begin{equation}
	\hat{\mathbf{g}}_t = [\hat{p}_{t,i}]_{i=0,1},
\end{equation}
\begin{equation}
	\hat{p}_{t,i} = \frac{\exp((\log(p_{t,i})+\epsilon_i)/\tau)}{\sum_{j=0}^1 \exp((\log(p_{t,j})+\epsilon_j)/\tau)},
\end{equation}
where $\epsilon_i$ is a random sample from Gumbel$(0,1)$. When temperature $\tau$ approaches $0$, Gumbel-Softmax distribution approaches to be one-hot.
The attention weights with soft gates can be computed by
\begin{equation}
\begin{aligned}
	\alpha_{t} &= \frac{g_t\odot\exp{e_t}}{\sum_{t'=1}^T g_{t'}\odot\exp{e_{t'}}}, \quad\sum_{t=1}^T \alpha_{t} = 1.
\end{aligned}
\end{equation}
We use the gradients of Gumbel-Softmax as the surrogate gradients during backpropagation.
During testing, however, the surrogate is not necessary. 
The generated gates are binary. Only selected elements are used for the computation of attention weights as in Eq(2).
And we directly use the probabilities generated by the auxiliary network to parameterize the Bernoulli distribution and obtain the binary gates. 

To facilitate training procedure, we define the loss of this joint network as follow:
\begin{equation}
	\mathcal{L} = -\sum_k y_k\log \hat{y}_k + \frac{\lambda\|\mathcal{G}\|_1}{T}.
\end{equation}
The first term in loss function is cross-entropy loss, where $y_k$ is the ground-truth label for $k$-th class.
The second term is an $l_1$ norm regularizer over all gates, where $\lambda$ is a hyper-parameter to make a trade-off between the cross-entropy loss and $l_1$ norm and $T$ is input sequence length. 
The $l_1$ norm term aims at encouraging the network to turn off more gates and generate more sparse attention connections.

\section{Experiments}
In this section, we evaluate the performance of GA-Net on various datasets for sentence classification tasks.
We did both quantitative and qualitative analysis on experiment results. 
Our model got better results compared with all baseline models in our experiments consistently. 

\begin{table}[t]
\caption{Datasets Statistics. This table provides average sequence length ($l$), number of classes ($K$), number of training samples (Train) and testing samples (Test), and type of task in each dataset.}\smallskip
\centering
\resizebox{\columnwidth}{!}{
\smallskip\begin{tabular}{l|c|c|c|c|c}
\toprule
Dataset & $l$ & $K$ & Train & Test & Types \\
\midrule
IMDB & 231 & 2 & 25k & 25k & sentiment analysis \\
AG & 44 & 4 & 120k & 7.6k & topic categorization \\
\midrule
SST-1 & 20 & 5 & 12k & 2.2k & sentiment analysis \\
SST-2 & 20 & 2 & 10k & 1.8k & sentiment analysis \\
TREC & 10 & 6 & 6k & 500 & question classification \\
\bottomrule
\end{tabular}
}
\label{table1}
\end{table}

\subsection{Text Classification}

\begin{table*}[t]
\caption{Results of classification accuracy on various datasets for text classification. Bold numbers indicate best performance. Numbers in brackets indicate density of attention connections.}\smallskip
\centering
\resizebox{.55\textwidth}{!}{
\smallskip\begin{tabular}{l|c|c|c|c|c}
\toprule
 					& IMDB		& AG's News	& SST-1		& SST-2		& TREC \\
\midrule
BiLSTM				& 0.8509 	& 0.9139 	& 0.4125 	& 0.8035		& 0.8876 \\
BiLSTM+localAtt 	& 0.8578 	& 0.9234 	& 0.4343 	& 0.8154 	& 0.8944 \\
BiLSTM+softAtt 		& 0.8863 	& \textbf{0.9264} 	& 0.4445 	& 0.8246 	& 0.9008 \\
\hline
GA-Net & \textbf{0.8941} & \textbf{0.9263} 	& \textbf{0.4464} & \textbf{0.8262} & \textbf{0.9124}\\
(density) & (0.1999) & (0.4310) & (0.4722) & (0.6005) & (0.4431)\\
\bottomrule
\end{tabular}
}
\label{table2}
\end{table*}

We ran a series of experiments on various datasets for sentence classification task. Table~\ref{table1} provides a summary of each dataset, and they are:
\begin{description}
\item[IMDB] Large Movie Review Dataset (IMDB) is a binary sentiment classification dataset. IMDB provides a set of 25,000 highly polar movie reviews for training, and 25,000 for testing.
\item[AG's News] AG's News corpus contains 496,835 categorized news articles from more than 2000 news sources. It is constructed into a topic classification dataset with 4 main categories by Xiang Zhang~\cite{zhang2015character}. Each category contains 30,000 training samples and 1,900 testing samples.
\item[SST-1] Stanford Sentiment Treebank is a collection of movie reviews but with train/dev/test splits provided and fine-grained labels (very positive, positive, neutral, negative, very negative), re-labeled by~\cite{socher2013recursive}\footnote{https://github.com/harvardnlp/sent-conv-torch/tree/master/data}. It has 11,855 training samples and 2,210 testing samples.
\item[SST-2] It is same as SST-1 but with neutral reviews removed and binary labels. It contains 9,613 training samples and 1,821 testing samples.
\item[TREC] This dataset is a collection of questions~\cite{li2002learning}. The task is to classify a question into 6 question types (whether the question is about entity, human, location information, etc.)
It contains 6,000 training samples and 500 testing samples.
\end{description}

For all the experiments, we chose a 2-layer bidirectional LSTM with attention as backbone network and another 1-layer bidirectional LSTM as auxiliary network (GA-Net). We applied the attention mechanism in~\cite{zhou2016attention} as benchmark for text classifications.
We compared our GA-Net with:
\begin{description}
	\item [BiLSTM] It is a vanilla bidirectional LSTM without attention. In this BiLSTM, we take the last hidden state as input to classifier; 
	\item [BiLSTM+localAtt] It is a vanilla bidirectional LSTM with local attention . The architecture we used as BiLSTM+localAtt is similar to that in~\cite{luong2015effective}. The original architecture in~\cite{luong2015effective} is designed for sequence-to-sequence model:  a local attended position in an encoder is predicted by the current hidden state in a decoder. To adapt it to our classification tasks, we take the last hidden state to predict a local position in a sequence to attend to.
	\item [BiLSTM+softAtt] This is a vanilla bidirectional LSTM with global soft attention. It is common used attention mechanism which computes an attention weight for each element in a sequence.
\end{description} 
In all three cases, we use the same BiLSTM configurations as those in backbone network of our GA-Net.

We download pre-trained 100-dimensional GloVe word vectors~\cite{pennington2014glove}\footnote{glove.6B, pretrained from Wikipedia 2014 and Gigaword 5. https://nlp.stanford.edu/projects/glove/} to initialize all models. 
The hidden dimensions of backbone BiLSTM and auxiliary BiLSTM are both 100 in our experiments.
During training, Gumbel-Softmax approximations are used as gates for all samples in both forward and backward propagation. 
During testing, only sampled discrete values are used. 
We use Adam~\cite{kingma2014adam} to optimize all models, with learning rate choosing from the set $[0.0001, 0.0002, 0.0005, 0.001, 0.002, 0.005, 0.01]$, batch size choosing from the set $[8, 16, 32, 64, 128]$. 

In GA-Net, we choose the temperature $\tau$ in Gumbel-Softmax from the set $[0.5, 1.0, 1.5, 2.0]$ 
to balance the sharpness of gradients and difficulty of training. We choose
$\lambda$ in loss function from the set $[0.4\times{10}^{-5},0.5\times{10}^{-5},1.0\times{10}^{-5}, 0.4\times{10}^{-4},0.5\times{10}^{-4},1.0\times{10}^{-4},1.0\times{10}^{-3}]$.
We adapt cross-validation to select hyper-parameters for each dataset and task. 

The results of classification accuracy are shown in Table~\ref{table2}. As it can be seen that, the proposed GA-Net consistently achieves the best performance on all datasets. For example, in TREC dataset, GA-Net achieves accuracy of 91.24\%, and outperforms the baselines by at least 1.16\%.
At the same time, for the density of the resulting attention connections, we can see that it achieves this with much sparser attention structures. Especially for long sequences as in IMDB dataset where the average sequence length is 231, only 19.99\% gates are switched on for each input. It demonstrates the fact that not all attention is needed especially for long sequences, and that GA-Net indeed has the ability of selecting those important units to attend to.
In BiLSTM+localAtt, we choose a window size of 40, 16, 8, 8, 4 for each dataset respectively. This aim at making BiLSTM+localAtt have similar sparsity with GA-Net and achieve good performance at the same time.

We also report the improvement of attention computation in GA-Net and BiLSTM+softAtt in Table~\ref{table3}. We measure the number of floating point operations (FLOPs) of attention computations during testing for both models. GA-Net with sparse attention has lower FLOPs than BiLSTM+softAtt.

\begin{table}[t]
\caption{The number of floating point operations of attention computations during testing (FLOPs).}\smallskip
\centering
\resizebox{0.99\columnwidth}{!}{
\smallskip\begin{tabular}{l|c|c|c|c|c}
\toprule
 					& IMDB		& AG's News	& SST-1		& SST-2		& TREC \\
\midrule
BiLSTM+softAtt 		& 2.4G 	& 131M 	& 17M 	& 14.1M 	& 1.4M \\
GA-Net & 0.4G & 59M 	& 6M & 5.6M & 0.7M \\
\bottomrule
\end{tabular}
}
\label{table3}
\end{table}

\begin{table}[t]
\caption{Case study in long sequences. The following are two paragraphs in IMDB test dataset. Bold texts are focused tokens selected by GA-Net.}\smallskip
\centering
\resizebox{\columnwidth}{!}{
\smallskip\begin{tabular}{l}
\toprule
\multicolumn{1}{p{10cm}}{\textbf{I} admire Deepa Mehta and \textbf{this movie is a masterpiece . I} 'd recommend to buy \textbf{this} movie on DVD because it 's a \textbf{movie you} might want to watch more often than \textbf{just} once . And trust me , you 'd still find little meaningful details after watching it several times.$<$br /$><$br /$>$ The \textbf{characters} - except for the grandmother perhaps - are all very balanced , no black and white . Even though you follow the story from the perspective of the two protagonists , there is also empathy for the other characters.$<$br /$><$br /$>$ I think the IMDb rating for the movie \textbf{is} far too low - probably due to \textbf{its politically controversial content} .}\\
\midrule
\multicolumn{1}{p{10cm}}{... means `` take up and read " , which is precisely what I felt like doing after having seen \textbf{this marvelous film.$<$br} /$><$br \textbf{/$>$Von} Ancken stimulates and inspires with \textbf{this breathtaking and superbly executed adaptation of} Tobias Wolff 's 1995 New Yorker article of the same name . The \textbf{incredible performance by} Tom Noonan \textbf{is brilliant and provocative and the editing} , sound design , cinematography and directing are \textbf{truly inspired . The nuanced} changes and embellishments on the original story are subtle , clever , and make the film cinematically more dynamic . It 's lyrical pacing \textbf{is mesmerizing and begs you to} watch it again.$<$br /$><$br /$>$Watch out for this young director ... he 's going places .} \\
\bottomrule
\end{tabular}
}
\label{table4}
\end{table}

\begin{figure*}[t]
\centering
\includegraphics[width=0.95\textwidth]{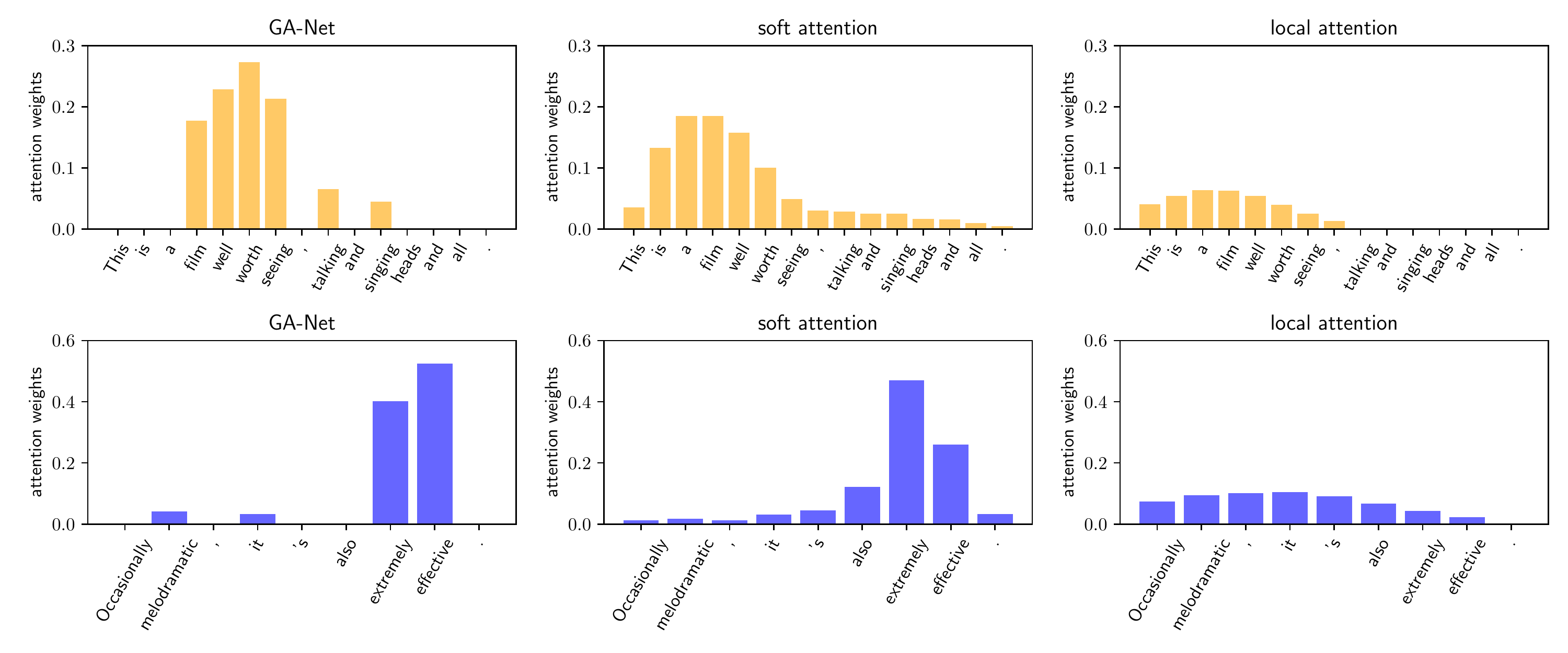}
\caption{Case study in short sequences. Each row is an example sentences with computed attention weights for each token. The first sentence is a positive statement: \emph{This is a film well worth seeing, talking and singing heads and all.} The second sentence is also a positive statement: \emph{Occasionally melodramatic, it 's also extremely effective.} The attentions from left column to right column are provided by GA-Net, soft attention and local attention respectively.}
\label{fig4}
\end{figure*}

\subsubsection{Interpretability} 
Since our GA-Net only attends to part of units in a given sequence, its capacity of selecting related units in the sequence is important. Though GA-Net has good performance, we still want to know what elements in a sequence they actually attend to. To explore this, we did several case studies. 

Figure~\ref{fig4} gives two examples. The two sentences in Figure~\ref{fig4} are drawn from SST-2 test dataset. Attention weights are computed for each token by GA-Net, soft attention and local attention respectively.
Attention weights in Figure~\ref{fig4}a and Figure~\ref{fig4}d are  assigned by GA-Net. The distributions of attention weights are sparse and compact. At the same time, attentions accurately focus on important token related to sentiment classification and shut off gates for meaningless tokens. For examples, token `\emph{extremely}' and `\emph{effective}' have very high weights in the second sentence ``\emph{Occasionally melodramatic, it 's also extremely effective.}''. Meaningless punctuations, token `\emph{'s}' and `\emph{Occasionally}' are shut off.
Figure~\ref{fig4}b and Figure~\ref{fig4}e are attentions computed by soft attention mechanism. Although it can identify related tokens, the distribution of its weights is more smooth than GA-Net and show less interpretability.
Local attention in Figure~\ref{fig4}c and Figure~\ref{fig4}f is much more smooth compared with the other two models. It exhibits a weak capacity of finding out related tokens. This also implies the reason why BiLSTM+localAtt has similar performance with raw BiLSTM especially for long sequences where attention is helpful.
Table~\ref{table4} are examples from IMDB dataset with attention computed by GA-Net. The bold texts are tokens being selected to attend. The attention results show that the GA-Net successfully identifies related keywords for classification.

\subsubsection{Auxiliary Network} 
\begin{figure*}[t]
\centering
\includegraphics[width=0.8\textwidth]{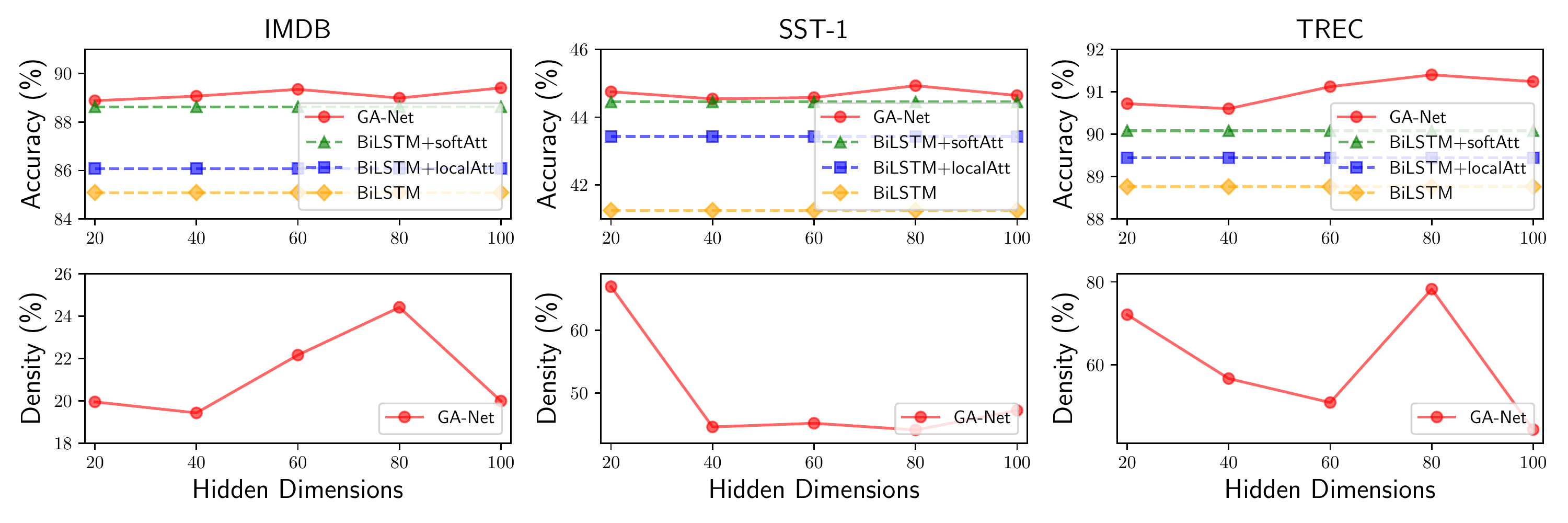} 
\caption{IMDB, SST-1 and TREC classification accuracy and attention density with different sizes of auxiliary networks. X-axis: size of LSTM hidden state; Y-axis: classification accuracy (upper row) and GA-Net attention density (bottom row). }
\label{fig5}
\end{figure*}

To investigate the impact from the size of hidden dimension in auxiliary network, we did several experiments on IMDB, SST-1, TREC
 with GA-Net using the same structure but with different hidden dimensions in auxiliary BiLSTM. The hidden dimensions range from 20 to 100 with a step of 20. Figure~\ref{fig5} provides the variations in accuracy and density.
we can observe that the performance of auxiliary LSTM with 20$d$ hidden states is competitive to that with 100$d$ hidden states. It also outperforms all baselines, and retains a low attention density.
This implies that just a small auxiliary network is able to make attention sparse.

\begin{table}[t]
\caption{Experiment results on IMDB Reviews with different auxiliary networks in GA-Net.}\smallskip
\centering
\smallskip\begin{tabular}{l|c c}
\toprule
GA-Net & Accuracy & Density\\
\midrule
BiLSTM+AUX$_{FNN}$ & 0.8890 & 0.2178 \\
BiLSTM+AUX$_{ATT}$ & 0.8892 & 0.5326 \\
BiLSTM+AUX$_{LSTM}$ & \textbf{0.8941} & \textbf{0.1999} \\
\bottomrule
\end{tabular}
\label{table5}
\end{table}

Moreover, we also curious about the impact from different choices of auxiliary networks. 
In stead of LSTM, we also chose an 1-hidden layer feed forward neural network (FNN), a self-attention network as auxiliary network to make comparison. We name them as BiLSTM+AUX$_{FNN}$, BiLSTM+AUX$_{ATT}$ respectively. We name the above GA-Net with LSTM as BiLSTM+AUX$_{LSTM}$. The backbone attention networks are same in all models. We use the same dimensions of hidden states as BiLSTM+AUX$_{LSTM}$ for BiLSTM+AUX$_{FNN}$ and BiLSTM+AUX$_{ATT}$. We did experiments on IMDB dataset.

Table~\ref{table5} gives the results. Both BiLSTM+AUX$_{FNN}$ and BiLSTM+AUX$_{ATT}$ outperform all baseline models. This proves again that not all attention is needed, and the ability of auxiliary network in selecting meaningful units.
Compared with BiLSTM+AUX$_{FNN}$ and BiLSTM+AUX$_{ATT}$, BiLSTM+AUX$_{LSTM}$ is still the one who achieves the best performance considering both accuracy and density.
Therefore, LSTM is a relatively good choice for auxiliary network in dealing with similar tasks in which the inputs are sequences.

\section{Conclusions}
In this paper, we propose a novel method called Gated Attention Network (GA-Net) for sequence data. GA-Net dynamically selects a subset of elements to attend to using an auxiliary network, and computes attention weights to aggregate the selected elements. It combines two input-dependent dynamic mechanisms, attention mechanism and dynamic network configuration, and has a dynamically sparse attention structure. Experiments show that the proposed method achieves the best results consistently while requiring less computation and achieving better interpretability. 

\section{Acknowledgments}
We would like to acknowledge the anonymous reviewers for their insightful comments. Research on this article was supported by Hong Kong Research Grants Council under grants 16202118 and 16212516.

\bibliographystyle{aaai}
\bibliography{AAAI-XueL.7235}

\end{document}